\documentclass[lettersize,journal]{IEEEtran}
\usepackage{amsmath,amsfonts}
\usepackage{algorithmic}
\usepackage{algorithm}
\usepackage{array}
\usepackage[caption=false,font=normalsize,labelfont=sf,textfont=sf]{subfig}
\usepackage{textcomp}
\usepackage{stfloats}
\usepackage{url}
\usepackage{verbatim}
\usepackage{graphicx}
\usepackage{amsfonts}
\usepackage{cite}
\begin{document}

\title{DSR-Diff: Depth Map Super-Resolution with Diffusion Model}

\author{Yuan Shi, Bin Xia, Rui Zhu, Qingmin Liao, and Wenming Yang
\thanks{This work was supported by the National Natural Science Foundation of China(Nos.62171251$\&$62311530100) and the Special Foundation for the Development of Strategic Emerging Industries of Shenzhen(JCYJ20200109143010272). (Corresponding author: Wenming Yang.)}
\thanks{Y. Shi, B. Xia, Q. Liao and W. Yang are with the Shenzhen International Graduate School, Tsinghua University, Shenzhen 518055, China(e-mail: shiy22@mails.tsinghua.edu.cn;zjbinxia@gmail.com; liaoqm@tsinghua.edu.cn; yang.wenming@sz.tsinghua.edu.cn).}
\thanks{R. Zhu is with Bayes Business School, City, University of London, London EC1V 0HB, United Kingdom(e-mail: rui.zhu@city.ac.uk).}
}

\markboth{Journal of \LaTeX\ Class Files,~Vol.~14, No.~8, August~2021}%
{Shell \MakeLowercase{\textit{et al.}}: A Sample Article Using IEEEtran.cls for IEEE Journals}

\IEEEpubid{0000--0000/00\$00.00~\copyright~2021 IEEE}

\maketitle

\begin{abstract}
  Color-guided depth map super-resolution (CDSR) improve the spatial resolution of a low-quality depth map with the corresponding high-quality color map, 
  benefiting various applications such as 3D reconstruction, virtual reality, and augmented reality.
  While conventional CDSR methods typically rely on convolutional neural networks or transformers, 
  diffusion models (DMs)  have demonstrated notable effectiveness in high-level vision tasks.
  In this work, we present a novel CDSR paradigm that utilizes a diffusion model within the latent space to generate guidance for depth map super-resolution.
  The proposed method comprises a guidance generation network (GGN), a depth map super-resolution network (DSRN), and a guidance recovery network (GRN). 
  The GGN is specifically designed to generate the guidance while managing its compactness. 
  Additionally, we integrate a simple but effective feature fusion module and a transformer-style feature extraction module into the DSRN, 
  enabling it to leverage guided priors in the extraction, fusion, and reconstruction of multi-model images.
  Taking into account both accuracy and efficiency, our proposed method has shown superior performance in extensive experiments when compared to state-of-the-art methods.
  Our codes will be made available at https://github.com/shiyuan7/DSR-Diff.
  \end{abstract}
  

  \IEEEpeerreviewmaketitle

  \section{Introduction}
  
  \IEEEPARstart{D}{epth} 
  information is a crucial element in understanding three-dimensional spaces, complementing color data and 
  playing a pivotal role in various computer vision applications, 
  such as 3D reconstruction\cite{choe2021volumefusion}, SLAM\cite{tateno2017cnn} and semantic segmentation\cite{weder2020routedfusion}.
  However, the practical use of modern depth sensors in consumer devices faces limitations due to budget and low-power constraints, 
  resulting in significantly lower resolutions compared to their RGB counterparts. 
  For example, the resolution of depth and color maps acquired by Kinect v2 is $512 \times  424 $ pixels and $1920\times 1080$ pixels, respectively.
  This resolution disparity hinders the effectiveness of the depth modality.
  \par
  Considering the accessibility of high-resolution color images, 
  various methods in color-guided depth super-resolution have emerged to harness both color and depth information. 
  Existing CDSR methodologies fall into three main categories: filtering-based methods, optimization-based methods, and learning-based methods. 
  Filtering-based methods pose a risk of introducing artifacts, as they may inadvertently transfer the texture of the guidance image to the depth map. 
  Meanwhile, optimization-based methods \cite{zuo2016explicit,xu2022depth} often involve time-intensive processes and may not yield satisfactory results when employing manually designed objective functions.
  \par
  Learning-based methods, which acquire the mapping between low-resolution (LR) and precise high-resolution (HR) depth maps, 
  have been proposed to address the limitations of hand-designed objective functions and filters \cite{zuo2019multi,zuo2019depth}. 
  Being at the forefront of employing a learning-based method to tackle the challenges posed by the CDSR problem, 
  MSG-Net\cite{hui2016depth} utilizes a CNN structure and employs multi-scale depth and color feature maps to enhance both the structure and texture of the HR depth maps.
  To leverage the edge information of the color maps, Wang et al.\cite{wang2020depth} proposed a two-stage method consisting of an edge-aware learning framework and depth restoration modules.
  Then, JIIF\cite{tang2021joint} was proposed to learn the interpolation weights, informed by the graph attention mechanism and implicit neural representation.
  Utilizing an efficient CNN architecture, AHMF\cite{zhong2021high} introduced an attention-based fusion mechanism for the integration of guidance information.
  
  \IEEEpubidadjcol
  
  \par
  While learning-based methodologies have shown significant efficacy, the performance of these models is ultimately limited by the inherent capabilities of foundational models like CNN or Transformer.
  Recently, diffusion models have exhibited pronounced effectiveness in tasks encompassing both the synthesis\cite{ho2020denoising,song2020denoising} and restoration\cite{saharia2022image,xia2023diffir} of images.
  Diffusion models can iteratively generate high-quality images from Gaussian white noise. 
  In comparison to CNNs and transformers, diffusion models exhibit stronger data fitting capabilities.
  \par
  Nevertheless, the utilization of diffusion models for depth map super-resolution is confronted with a series of impediments.
  Firstly, the intrinsic iterative nature of diffusion models results in a temporal protraction.
  Secondly, the potent generative capability of diffusion models can lead to artifacts.
  To address these issues, we adopt a two-stage training approach and implement diffusion models in latent space, 
  drawing inspiration from \cite{xia2023diffir}, to circumvent the need for recovering the entire image in CDSR problem.
  Specifically, in the first stage, we compress the ground truth depth map, color image and low-resolution depth map to generate a guiding prior using the proposed Guidance Generation Network (GGN).
  And an effective and efficient depth map super-resolution network(DSRN) is designed to utilize the guiding prior and reconstruct the depth map.
  Then the guiding prior is generated by guidance recovery network(GRN) with low-resolution depth map, relying on the powerful generative capability of diffusion models.
  Generally, the main contributions of our work are made as follows:
  
  $\bullet  $ We present a novel method, DSR-Diff, as the pioneering effort in applying diffusion models in latent space to enhance the resolution of depth maps, yielding noteworthy results.
  
  $\bullet  $ We devised the GGN to systematically generate compressed guiding priors. To fully harness these guiding priors, we introduced the DSRN, which integrates a guidance embedding module with our proposed simple yet efficient feature fusion module.
  
  $\bullet  $ Extensive experiments show that our proposed method achieves state-of-the-art performance, considering both precision and efficiency.
  
  \begin{figure*}[!htb]
     \centering
     \includegraphics[width=0.95\linewidth]{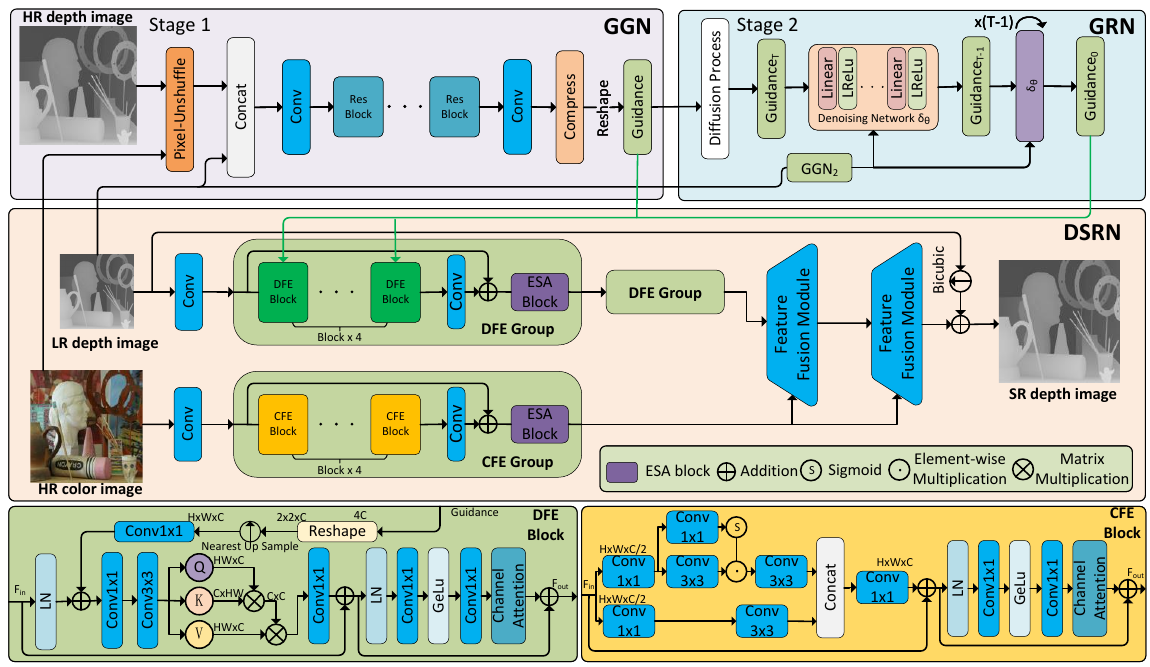}\\
     \caption{The architecture of the proposed DSR-Diff.
     (a) DSRN utilizes two branches to extract features from LR depth images and HR color images, followed by feature fusion and reconstruction of the depth map, respectively.
     (b) In the first stage, LR depth images, HR depth images, and HR color images are input into GGN to generate guidance, directing the feature extraction and reconstruction in DSRN.
     (c) In the second stage, GRN is designed to estimate the guidance generated by GGN, thanks to the powerful estimation abilities of the diffusion model.
     }
     \label{fig1}
  \end{figure*}
  
  \section{Proposed Method}
  \subsection{Overview}
  Figure \ref{fig1} shows our overall network architecture. We train DSR-Diff in two stages. 
  In the proposed framework, DSRN functions as the main restoration network, employing LR depth images, HR color images, and guidance as inputs to reconstruct SR depth images.
 During the first stage, GGN generates guidance by extracting and compressing valuable features from HR color images, HR depth images, and LR depth images.
 In the second stage, GRN generates guidance. In the forward process, the diffusion model progressively introduces noise into the guidance produced in the first stage. During the reverse process, the diffusion model utilizes LR depth images to estimate the necessary guidance.

  \subsection{Guidance Generation Network(GGN)}
  To provide accurate and concise guidance for DSRN, GGN is designed to extract essential information from HR depth images, HR color images, and LR depth images in the first stage of training.
  HR depth and color images are initially downsampled using pixel-unshuffle according to the scale factor.
  Subsequently, the downsampled images and LR depth images are concatenated and passed through several resblocks to generate guidance information.
  The feature maps of each channel in the guidance are divided into $K \times K $ blocks, and then we employ an operation similar to pooling to compress the information within these blocks.
  Given an concatenated feature $G\in \mathbb{R} ^{H\times W\times C}$, the compressed guidance is $\hat{G}\in \mathbb{R} ^{K^{2} \times  C}$. We set $K = 2$ in this problem.

  \subsection{Guidance Recovery Network(GRN)}
  In the second stage, we use GRN to generate guidance from the LR depth image. 
  Guidance $G \in \mathbb{R} ^{K^{2}C}$ is initially generated by the GGN trained in the first stage. And we apply the diffusion process on $G$ to sample $G_{T}$, which can be described as:
  \begin{equation}\label{eq2}
     q(G_{T}\vert G) = \mathcal{N} (G_{T}; \sqrt{\overline{\alpha }_{T}}G, (1-\overline{\alpha }_{T} )\mathcal{I}  ),
  \end{equation}
  where T is the total number of iterations, $\mathcal{N}()$ denotes the Gaussian distribution.
  \par
  Since the guidance required for DSRN is very simple and compact, we can use a straightforward denoising network to estimate noise, making the guidance estimation simple and efficient.
  We start from $T_{th}$ time step and run denoising iterations to obtain the estimated $G_{0}$ and send it to DSRN for joint optimization,
  which can be formulated as:
  \begin{equation}\label{eq3}
     \hat{G}_{t-1} = \frac{1}{\sqrt{\alpha _{t}} }(\hat{G}_{t}- \epsilon \frac{1-\alpha _{t}}{\sqrt{1-\overline{\alpha }_{t} } } )  ,
  \end{equation}
  where $\epsilon $ is the same noise. And in the reverse process, the LR depth images are initially input into $GGN_{2}$, which has the almost same architecture with GGN except for the input, to generate the condition $C_{G}$.
  Then $C_{G}$, $G_{t}$ and t are input to the denoising network $\delta _{\theta }$ to estimate noise. After T times iterations, we get the estimated guidance $G_{0} \in \mathbb{R} ^{K^{2}C}$.

  \subsection{Depth Map Super-Resolution Network(DSRN)}
  DSRN can utilize the guidance generated by GGN or GRN to guide the feature extraction of depth images and obtain the reconstructed SR depth maps.
  DSRN primarily consists of the depth image feature extraction group (DFE group), the color image feature extraction group (CFE group), and the feature fusion module.
  In our pursuit of reconstructing depth images with a focus on both efficiency and accuracy, we opt for the utilization of two DFE groups to process depth features, alongside a dedicated CFE group specifically designed for handling color features.
  \par
  By concatenating multiple DFE Blocks and incorporating an Enhanced Spatial Attention (ESA) block \cite{liu2020residual} in a sequential manner, 
  a feature extraction architecture, denoted as the DFE Group, is obtained. 
  We devised the DFE block based on the transformer block in \cite{zamir2022restormer}. 
  And to leverage the guidance, the input guidance is first reshaped, upsampled to match the size of the feature map, 
  and then added to the feature map after a 1x1 convolution. Given an input feature $F_{in} \in \mathbb{R} ^{H\times W\times C}$ and guidance $G \in \mathbb{R} ^{K^{2}C}$,
  This process can be formulated as:
  \begin{equation}\label{eq1}
    \hat{F} = Conv(Up(Reshape(G))) + LN(F_{in}),
   \end{equation}
  where $\hat{F}\in \mathbb{R} ^{H\times W\times C}$ is the guidance-adjusted feature map, $Reshape(\cdot )$ is the reshape operation and $Reshape(G) \in \mathbb{R} ^{K\times K\times C}$, 
  $K$ is a hyperparameter that we can use to control the compactness of guidance, $UP(\cdot )$ is the nearest upsampling, 
  $Conv(\cdot )$  refers to a convolution with a kernel size of 1, and $LN(\cdot )$ is the layer normalization.
  \par
  The structure of the CFE Group is consistent with that of the DFE Group.
  Due to the fact that the color image feature extraction branch does not require parsing information from the guidance, 
  to enhance the efficiency of the model, we employed a simpler approach to design the CFE block inspired by \cite{zhao2020efficient} as shown in figure \ref{fig1}.
  
  \begin{figure}[!htb]
     \centering
     \includegraphics[width=\linewidth]{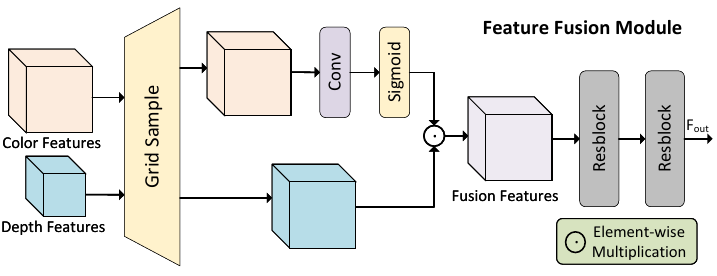}\\
     \caption{The architecture of our proposed feature fusion module (FFM).
     }
     \label{fig2}
  \end{figure}
  \par
  
  The extracted deep features and color features are upsampled to the same size and then fused within the feature fusion module (FFM), as illustrated in figure \ref{fig2}. 
  The reconstructed SR depth image is obtained after passing through two FFMs.
  Depth and color features are upsampled through grid sampling, allowing the sampling of features at specific coordinates, thereby mitigating the impact of positional offsets in depth and color features.
  \par
  Traditional learning-based DSR methods such as \cite{tang2021joint} and \cite{hui2016depth} often directly concatenate the color and depth features
  which tend to introduce unnecessary features in color image and require additional computation resources.
  In our proposed FFM, convolution is first applied to the complex color features, and then the upsampled color and depth features are directly multiplied together,
  which allows the effective components for reconstruction in both depth and color features to be retained, while weakening other non-essential features.
  Furthermore, two resblocks are concatenated at the end of the FFM for the refinement of the fused features.
  
  \subsection{Loss Function}
  In the first stage, our objective is to generate an effective guidance and get high-quality depth map through DSRN. 
  So the training loss is defined as follows:
  \begin{equation}\label{loss1}
     L_{img} = \| D_{GT} - D_{SR} \|_{1} 
  \end{equation}
  where $D_{GT}$ is the ground-truth depth image, $D_{SR}$ is the reconstructed depth image, and $\| \cdot \|_{1} $ denotes the $L_{1}$ norm.
  \par
  In the second stage, we need to estimate the guidance in the first stage and reconstruct the SR depth image. So we joint train the DRN and DSRN using loss function $L_{com}$:
  \begin{equation}\label{loss2}
     L_{com} = \| G - G_{0} \|_{1} + \| D_{GT} - D_{SR} \|_{1} 
  \end{equation}
  And after training the model for a specified number of epochs, we still use $L_{img}$ as the loss function for better reconstruction quality.

  \section{Experiments}
  \label{sec:guidelines}
  
  \subsection{Implementation Details}
  
  \subsubsection{Dataset}
  We select three benchmark RGB-D datasets, NYU V2 dataset \cite{silberman2012indoor}, 
  Middlebury dataset \cite{hirschmuller2007evaluation,scharstein2007learning}, and Lu dataset \cite{lu2014depth}, to evaluate our method.
  NYU V2 dataset includes 1449 RGB-D image pairs, with 1000 image pairs for training and the remaining 449 pairs for evaluation.
  Middlebury dataset and Lu dataset each have 30 and 6 RGB-D images pairs, respectively, all of which are utilized for evaluation.

  \subsubsection{Training Setting}
  We apply a batch-mode learning method with a batch size of 1.
  Both the first and second stages require training for 300 epochs each.
  And The initial learning rate is set to $2 \times 10^{-4}$ and decays by half every 80 epochs.
  The Adam optimizer with $\beta _{1}=0.9$, $\beta _{2}=0.99$ are used for optimization. 
  In training the diffusion model, time step $T$ are set to 4 and $\beta _{t}$ ($\alpha _{t}=1-\beta _{t}$) linearly increase from 0.1 to 0.99.
  And all the experiments are performed with PyTorch on PC with one GTX 3090 GPU.
   \begin{table*}[!htb]
      \centering
      \caption{Quantitative comparison with state-of-the-art methods on NYU v2, Middlebury, and Lu datasets in terms of RMSE (the lower, the better). The best performance is shown in \textbf{blod}, while the second best performance is the \underline{underscored} ones.}
      \label{sota1}
      \resizebox{0.72\textwidth}{!}{
      \begin{tabular}{cccccccccc}
          \hline  
         Method  &\multicolumn{3}{c}{NYU V2} &   \multicolumn{3}{c}{Middlebury}  &\multicolumn{3}{c}{Lu}\\ \cline{2-10}
            & $4\times$    &$8\times$   & $16\times$    &$4\times$& $8\times$  &$16\times$   &$4\times$& $8\times$  &$16\times$  \\ \hline
          Bicubic&4.28&7.14& 11.58 &2.28 & 3.98 &6.37  &2.42  &4.54  &7.38     \\\cline{1-10}

          MSG-Net \cite{hui2016depth} &3.02&5.38    &9.17 &1.88  &3.45   &6.28 &2.30  &4.17   &7.22  \\
          DJF \cite{li2016deep} &3.54 &6.20 &10.21 &2.14 &3.77 & 6.12 & 2.54 & 4.71&7.66 \\
          DG \cite{gu2017learning}  &3.68 &5.78 &10.08 &1.97 &4.16 & 5.27 & 2.06 &4.19 & 6.90 \\
          DJFR \cite{li2019joint} &2.38&4.94&9.18&1.32&3.19&5.57&1.15&3.57&6.77\\
          CUNet \cite{deng2020deep} &1.92&3.70    &6.78 &1.10  &2.17  &4.33 &0.91  &2.23   &5.19  \\
          FDKN \cite{kim2021deformable} &1.86&3.55    &6.96&1.09  &2.17   &4.51 &\textbf{0.82}   &2.09   &5.03 \\
          DKN \cite{kim2021deformable} &1.62&3.26    &6.51 &1.23  &2.12   &4.24 &0.96   &2.16   &5.11  \\
          DCTNet \cite{zhao2022discrete}  &1.59&3.16 &5.84 &1.10  &2.05   &4.19 &0.88   &1.85   &4.39  \\
          JIIF\cite{tang2021joint} &\underline{1.37} & \underline{2.76}    &\underline{5.27} &1.09  &1.82   &3.31 & \underline{0.85}   &1.73   &4.16  \\
          AHMF\cite{zhong2021high} &1.40 &2.89    &5.64 &\underline{1.07}  & \textbf{1.63}   &\underline{3.14} &0.88   &\underline{1.66}   & \textbf{3.71}  \\
          Ours &\textbf{1.25}&\textbf{2.57}    &\textbf{4.91} &\textbf{1.04}  &\underline{1.67}   &\textbf{3.10} &0.91    &\textbf{1.65}   &\underline{3.92} \\ \hline
      \end{tabular}}
    \end{table*}
    
    \begin{figure*}[!htb]
       \centering
       \includegraphics[width=\linewidth]{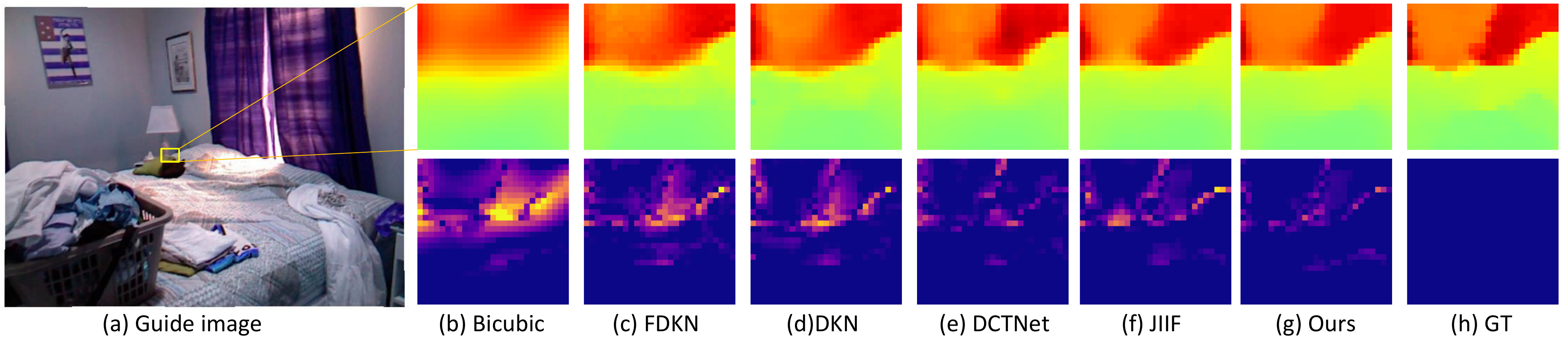}\\
       \caption{Qualitative comparisons for ×8 CDSR on the NYU v2 dataset.The first row is a zoomed-in view of the reconstructed depth map.
       The second row displays the error map between the results and the ground truth and brighter areas indicate more significant errors.
       }
       \label{visual1}
    \end{figure*}

  \subsection{Comparisons with State-of-the-Art Methods}
  We compare DSR-Diff with state-of-the-art methods.
  Quantitative comparison results at scales of 4x, 8x, and 16x are presented in Table \ref{sota1}. We use the average RMSE as the evaluation metric.
  Following \cite{tang2021joint}, 
   the average RMSE is measured in centimeters for the NYU v2 dataset. 
   For the other two datasets, RMSE computations are carried out after scaling the depth values within the [0, 255] range.
   \par
   \begin{figure}[!htb]
      \centering
      \includegraphics[width=0.95\linewidth]{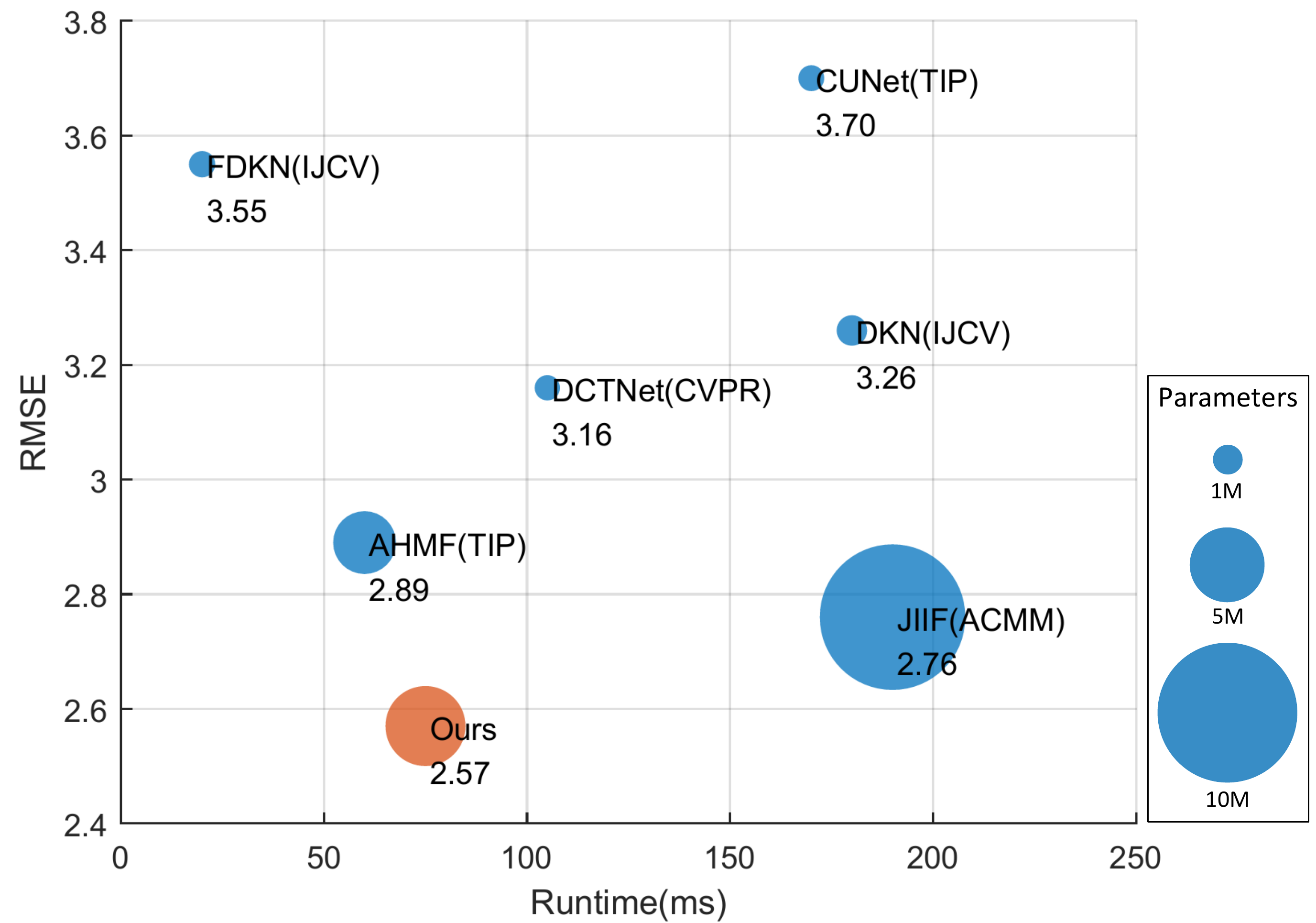}\\
      \caption{Efficiency comparison with the current state-of-the-art methods for 8x CDSR on the NYU v2 dataset. 
      The RMSE results are indicated below the name of each method, and the circular area serves as a visual representation of the model's parameters.}
      \label{com1}
   \end{figure}

   \par
   The results in Table \ref{sota1} clearly indicate that our proposed methods outperform alternative approaches in terms of accuracy.
   This superiority is particularly evident in the NYU v2 and Middlebury datasets, both of which include more test images.
   To further illustrate the superiority of our proposed method, 
   we compared its runtime, parameters, and accuracy with the current state-of-the-art methods, as depicted in Figure \ref{com1}.
   It is evident that, considering both efficiency and accuracy, our proposed method achieves state-of-the-art performance.
   \par
   To better demonstrate the performance of DSR-Diff, we provide an intuitive visual comparison in Fig. \ref{visual1}.
   It can be clearly seen that the depth images reconstructed by DSR-Diff exhibit superior visual results, especially in areas with complex textures and edges.

  \subsection{Ablation Studies}
  To validate the effectiveness of each module in DSR-Diff, we conducted ablation experiments and the results are presented in Table \ref{ab1}.
  \begin{table}[!htb]
   \centering
   \caption{Ablation studies on the NYU v2 test set}
   \label{ab1}
   \resizebox{0.9\columnwidth}{!}{
   \begin{tabular}{cccccc}
       \hline  
       & M-1\centering&M-2\centering  &M-3 \centering &  M-4  \centering&  M-5   \\ \hline
       GGN &$\times$\centering  &$\times$ \centering  & $\surd$ \centering  & $\surd$\centering  & $\surd$      \\
       GRN &$\times$ \centering &$\surd$ \centering   & $\times$ \centering  & $\surd$  \centering  & $\surd$    \\
       FFM &$\surd$ \centering &$\surd$ \centering &$\surd$ \centering & $\times$\centering  & $\surd$      \\ \hline
       Average RMSE&  2.643\centering  & 2.597 \centering   & 2.570 \centering&2.587   \centering&2.572     \\ \hline
   \end{tabular}}
 \end{table}
 \par
In M-1, the guidance is not incorporated into DSRN and we solely utilize the designed DSRN for depth image reconstruction.
In M-2, the compress operation in GGN is replaced with a regular pooling operation. 
M-3 presents the results of the first stage. In conjunction with M-5, it substantiates the efficacy of the diffusion model in the second stage, aligning with the guidance generated by GGN.
In M-4, the upsampled depth features and color features in FFM are directly concatenated together.
The effectiveness of our proposed modules in enhancing the model's performance is clearly evident.

  \section{Conclusion}
  In this paper, we propose DSR-Diff, a guided super-resolution method with diffusion models. 
  To address the time-consuming nature of diffusion models, we apply diffusion models in the latent space to generate compact guidance.
  Furthermore, we jointly optimize the diffusion model with an effective and efficient depth map super-resolution network (DSRN) to leverage the generated guidance.
  And a guidance generated network (GGN) is proposed to ensure that the guidance is appropriately generated and compressed.
   The feature fusion module (FFM) employs a simple yet efficient approach to achieve the integration of multi-modal features.
   Extensive experiments validate the effectiveness and superiority of our proposed DSR-Diff.

\bibliographystyle{IEEEtran}
\bibliography{Template}

\vfill

\end{document}